\documentclass{article}

\usepackage{PRIMEarxiv}

\usepackage[utf8]{inputenc} % allow utf-8 input
\usepackage[T1]{fontenc}    % use 8-bit T1 fonts
\usepackage{float}
\usepackage{hyperref}       % hyperlinks
\usepackage{url}            % simple URL typesetting
\usepackage{booktabs}       % professional-quality tables
\usepackage{amsfonts}       % blackboard math symbols
\usepackage{nicefrac}       % compact symbols for 1/2, etc.
\usepackage{microtype}      % microtypography
\usepackage{lipsum}
\usepackage{fancyhdr}       % header
\usepackage{graphicx}       % graphics
\usepackage{amsmath}
\graphicspath{{media/}}     % organize your images and other figures under media/ folder
\usepackage[table]{xcolor}
\usepackage{adjustbox}  
\usepackage{natbib}

\usepackage[scaled]{helvet} % for sans-serif text
 
\usepackage{mathpazo} % for Palatino-like math fonts
\title{Class-Conditional self-reward mechanism for improved Text-to-Image models}
\author{
  Safouane EL GHAZOUALI*\\
  TOELT LLC AI lab\\
  Winterthur, Swintzerland \\
  \texttt{safouane.elghazouali@toelt.ai} \\
  \And
  Arnaud GUCCIARDI\\
  TOELT LLC AI lab, \\
  Winterthur, Swintzerland \\
  \texttt{arnaud.gucciardi@toelt.ai}\\
  \And
  Umberto Michelucci \\
  TOELT LLC AI lab, \\
  Winterthur, Swintzerland \\
  \texttt{umberto.michelucci@toelt.ai} \\
}

\begin{document}
\maketitle

\begin{abstract}
  Self-rewarding have emerged recently as a powerful tool in the field of Natural Language Processing (NLP), allowing language models to generate high-quality relevant responses by providing their own rewards during training. This innovative technique addresses the limitations of other methods that rely on human preferences. In this paper, we build upon the concept of self-rewarding models and introduce its vision equivalent for Text-to-Image generative AI models. This approach works by fine-tuning diffusion model on a self-generated self-judged dataset, making the fine-tuning more automated and with better data quality. The proposed mechanism makes use of other pre-trained models such as vocabulary based-object detection, image captioning and is conditioned by the a set of object for which the user might need to improve generated data quality. The approach has been implemented, fine-tuned and evaluated on stable diffusion and has led to a performance that has been evaluated to be at least 60\% better than existing commercial and research Text-to-image models. Additionally, the built self-rewarding mechanism allowed a fully automated generation of images, while increasing the visual quality of the generated images and also more efficient following of prompt instructions. The code used in this work is freely available on \href{https://github.com/safouaneelg/SRT2I}{GitHub}.
\end{abstract}

\section{Introduction}
Self-rewarding is a concept that has been recently introduced in the field of Natural Language Processing (NLP) by [\cite{yuan2024selfrewarding}] to address the limitations of traditional reward modeling based on reinforcement learning. More specifically, in Large Language Models (LLM), the reward function is typically defined by human annotators who rate the quality of generated text based on their preferences (Reinforcement learning from human feedback RLHF) [\cite{ouyang2022training}]. This approach has several limitations, including the need for large amounts of human-annotated seed data, the potential for reward hacking (where the model learns to manipulate the reward function to generate higher-rewarding outputs), and the difficulty of accurately capturing human preferences in a reward function.

Therefore, the self-reward pipeline proposed in [\cite{yuan2024selfrewarding}] has the object of self-improving the LLM continuously through a model that can, not only generate text, but also self-judge its own content and improving the quality of the data. The authors suggest the LLM self-reward pipeline as follow: (1) start by generating a set of prompts using a pretrained model. (2) following this, N candidate responses are generated for each specific prompts by the same LLM. (3) the chosen responses is selected along with their prompts as 'preference pairs'. (4) the model is then trained using Direct Preference Optimization (DPO). DPO is an LLM training approach that has been proposed recently by Rafailov \textit{et. al.} [\cite{rafailov2023direct}]. This method aims to align LLMs with human preferences by using a simple classification loss. DPO eliminates the need for sampling from the language model during the fine-tuning process.
The main idea behind DPO is to optimize the likelihood of the preferred response over the dis-preferred response [\cite{amini2024direct}]. This is done by establishing the preferred and dis-preferred answers to the model for a specific prompt. The approach leads to an optimized likelihood of the preferred response and therefore resulting in a policy reformulation. DPO is simple, compatible with various language models architectures, and, unlike other LLM training approaches, does not require reinforcement learning or complex reward modeling.

This paper explores the potential implementations of a concept analogous to self-improvement within the field of Text-to-Image (T2I) models. The overarching objective remains centered around similar rewarding concept to [\cite{yuan2024selfrewarding}], wherein the T2I model not only refines its performance but actively engages in a self-improvement process. This transformative framework is characterized by a process of self-judgment and autonomous decision-making capabilities, introducing a novel approach to model evolution.

Text-2-Image models, also known as diffusion models, are a class of generative models that learn to produce new images by simulating a diffusive process [\cite{zhang2023texttoimage}]. In the context of machine learning, this typically involves gradually adding noise to a data sample until the full image is covered with Gaussian Noise and then learning to reverse this process to generate new samples from the noise. This reverse generative process involves learning to transform the noisy samples into iteratively less and less noisy data. This can be performed using a neural network architecture that uses the information about diffused data and the current time step as inputs to output a prediction of the original data sample.

Drawing from a comprehensive understanding of the mechanics inherent to Text-2-Image (T2I) models, the exploration shifts towards the application of a self-rewarding approach specifically crafted for T2I models. For given prompts, the proposed T2I self-rewarding approach could be summarized in a 3-steps loop:% (Figure  \ref{fig:self-rewarding_loop}):

%\begin{figure}[h]
%	\centering
%        \includegraphics[width=80mm,scale=1]{self-rewarding_loop.pdf}
%	\caption{Suggested self-rewarding loop for Text-to-Image diffusion models.}
%	\label{fig:self-rewarding_loop}
%\end{figure}

\begin{itemize} \itemsep -2pt
    \item[$(1)$] For each prompt, multiple images are generated in order to simulate multiple case in which one will be the closest to the prompt.
    \item[$(2)$] Unlike LLM self-rewarding, T2I models are limited to image generation, therefore cannot be used as self-judge to their own data. We suggest to use the inverse models Image-to-Text (I2T) to identify each of the generated images view and scene detailed descriptions.
    \item[$(3)$] Out of those images, the optimal pairs of prompt-image will be retrieved and subsequently used to fine-tune the T2I model.
\end{itemize}

Within this work, we introduce a novel architecture-agnostic self-rewarding loop that can be seamlessly integrated with any T2I model. This versatile framework offers a new solution to enhance the accuracy of scene generation, enabling models to more effectively follow instructions and produce specific scenes featuring a given set of objects. By leveraging this approach, an improvements in model performance have been reached in text-to-image synthesis capabilities.

Therefore, the approach could be summarized as follow:
(1) Building a novel mechanism named 'class conditional self-rewarding' that allows the autonomous optimization of T2I models. (2) Integration of I2T model to the pipeline allowing image judging based on visual content. (3) Use of Open vocabulary object detection to track the accurate detection the initial object class. (4) Establishment of a selection policy for optimal pairs text - best describing image. (5) Fine-tuning of T2I models based on those selected pairs.

The key contributions of this work can be distilled into the following main points: Firstly, a novel mechanism called 'class-conditional self-rewarding' has been developed, allowing the autonomous optimization of T2I models for specific visual content. This mechanism not only enhances model performance but also boosts adaptability in various contexts. Secondly, incorporating I2T models into the pipeline offers an additional assessment based on captioning and visual content description, which allows for more nuanced judgments. Thirdly, the use of Open-vocabulary object detection techniques leads to enhanced precision, ensuring the accurate identification of initial object classes. Furthermore, a selection policy has been established to curate optimal text-image pairs, thereby enriching the descriptive capacity of the models. Finally, the fine-tuning of T2I models based on these selected pairs allowing efficacy and coherence in image interpretation and description.

The rest of this paper will be structured as follow: section \ref{sec:srt2i_theroy} presents the building of T2I self-rewarding pipeline while detailing the different steps. Section \ref{sec:Experimentalstudy} presents an experimental implementation when testing the self-rewarding on pre-trained models.

\section{Method: Class-conditional Self-rewarding}\label{sec:srt2i_theroy}

\begin{figure}[ht]
	\centering
        \includegraphics[width=\textwidth,scale=1]{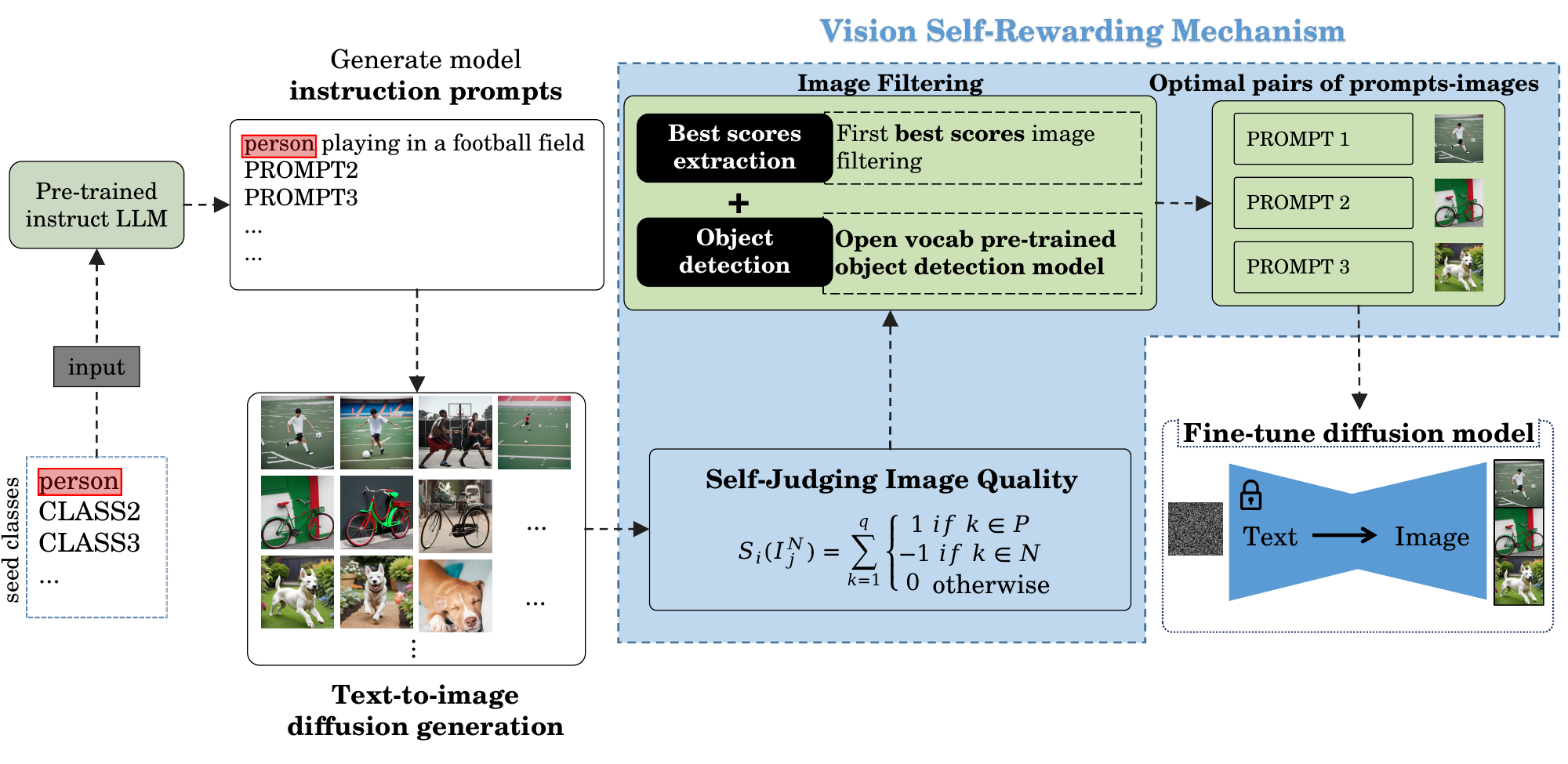}
	\caption{Overview flowchart of self-rewarding mechanism for Text-2-image models. Self-rewarding mechanism groups 3 steps: (1) Self-judging function, (2) Image filtering, (3) optimal pairs extraction.}
	\label{fig:flowchart_of_t2i_model}
\end{figure}

The proposed self-rewarding, initially inspired from [\cite{yuan2024selfrewarding}], works by combining the power of language performance and understanding of a LLM, text-to-image generation of image, and Image captioning also known as Image-to-text (I2T) to judge the generated images and provide the best possible dataset to T2I models. The self-rewarding pipeline could be summarized into 6 steps as shown in Figure \ref{fig:flowchart_of_t2i_model}.

The presented steps in the flowchart could be grouped into four actions: 
\begin{itemize}
    \item \textbf{T2I Prompts generation}:
    This groups the seeding of one or multiple specific classes to which we would like to the T2I model to be more optimal in image generation. The LLM is used as a multi-cases diverse scenarios descriptor.
    
    \item \textbf{Multi-Image construction}:
    Based-on the obtained prompts for the specified classes, those are going to be fed to the T2I model such to generate N candidate images for each specific scenario.
    
    \item \textbf{Self-rewarding mechanism}:
    The self-rewarding mechanism is used to filter out of those image the most corresponding one to the initial prompt. Additional optional steps are suggested to improve the accuracy of optimal pairs selection.
    
    \item \textbf{Fine-tuning}:
    The estimated optimal pairs based on the self-rewarding mechanism  are then used as input for the fine-tuning of the T2I model.

\end{itemize}
%Let's dive into more explanatory details about each specific action within the self-rewarding T2I flowchart (Figure \ref{fig:flowchart_of_t2i_model}). 

\subsection{T2I Prompts generation}
Existing Language Model (LLM) architectures have demonstrated quiet impressive capabilities in understanding human language and even making contexts and emotions interaction [\cite{li2023large}], making them usable for various automation tasks [\cite{zhang2023responsible, 10109345}]. Initially trained on massive textual datasets, those models have impressive text generation and completion capabilities and garnered attention for their proficiency in understanding nuanced language structures [\cite{wu2023survey}]. In the context of our research on Text-to-Image (T2I) diffusion models, leveraging the strengths of LLMs allows automating the generation of prompts.

To initiate the T2I Prompts generation process, the selection of an instruction-based fine-tuned LLM is required. Many LLM models have been fine-tuned for Q\&A chat such as Llama 2 [\cite{touvron2023llama}], Falcon series [\cite{almazrouei2023falcon}] and more recently Mistral [\cite{jiang2023mistral}] which seems to be efficient among low parameters models. Additionally, the advent of quantization allowed the efficient use of those models even on CPUs and low memory GPUs such as Activation-aware Weight Quantization (AWQ) [\cite{lin2023awq}], Generative Pre-trained Transformers Quantization (GPTQ) [\cite{frantar2023gptq}]. The selection of those quantized models is sufficient for the aforementioned process consisting of generating a list of prompts.

As the LLMs are usually trained on massive amounts of data, those might have limitless generation to a specific prompt. For the self-rewarding mechanism, we condition the prompts generation by one or multiple sets of object classes to limit its text generation making it suggesting diverse scenarios for those objects only.

\subsection{Text-to-Image MultiImage generation}
T2I model have the capacity to understand the overall request and tries to reconstruct an image based on the keyword in the prompt. Therefore, there are many possibilities of prompt-based image generation for a each single prompt. One way of limiting this to a defined image amounts is seeding the algorithm with one or multiple sets of random numbers. However the seeding limits also the creativity of the model since the goal is to make the T2I reproducible.

Similarly to the proposed "self-rewarding language model" framework, instead of seeding the T2I model, we propose the generation of N candidate images $\{\mathrm{I}_{0}^{N}, \mathrm{I}_{1}^{N}, ..., \mathrm{I}_{m}^{N}\}$ for each prompt $P_{i = (1,... m)}$.

\subsection{Self-Rewarding Mechanism}

The idea behind the proposed self-rewarding mechanism is to perform a fully automated and optimal selection of pairs $P_{i}$ - $\mathrm{I}_{j}^{N}$. To achieve this goal, the T2I model is given enough freedom for creativity by generating multiple images for each single prompt. Our hypothesis suggests that at least one of those images will be the closest to the given prompt, and therefore considered as the optimal image for the pair $P_{i}$ - $\mathrm{I}_{j}^{N}$. %(Figure \ref{fig:self_reward_from_multi-images}).

%\begin{figure}[ht]
%	\centering
%        \includegraphics[width=100mm,scale=1]{self-rewarding-from_multiimage.pdf}
%	\caption{visualization of the self-reward policy function from multi-images generation. Where (1) is the first path for image generation and (2) is the second path of optimal pairs extraction}
%	\label{fig:self_reward_from_multi-images}
%\end{figure}

The process involves extracting images from a set of N generated ones $I_{j}^{N}$ using T2I model, resulting in a set of images denoted as  for a given prompt $P_{i}$. The generated set of N images $I_{j}^{N}$ are subjected to a collection of evaluative questions $Q_{k=1,..,q}$, and the assistant provides binary answers ('yes' or 'no') to each question. The scoring mechanism, denoted by $S_{i}$ for each image, is determined based on the answers provided.

Let $Q_{k}$ represent the set of evaluative questions, and $A_{k}$ represent the corresponding answers provided by the assistant. The scoring function $S_{i}$ for image $I_{j}^{N}$ is defined as follows:

\begin{equation}
    S_{i}(I_{j}^{N}) = \sum_{k=1}^{q} \left\{
    \begin{array}{ll}
        1 & \text{if } (k \in \mathcal{P} \text{ and } A_{k} = \text{'yes'}) \text{ or } (k \in \mathcal{N} \text{ and } A_{k} = \text{'no'}) \\
        -1 & \text{if } k \in \mathcal{N} \text{ and } A_{k} = \text{'yes'} \\
        0 & \text{otherwise}
    \end{array}
    \right.
\end{equation}

$\mathcal{P}$ represents the set of indices corresponding to positive questions, and $\mathcal{N}$ is the set of indices corresponding to negative questions. The domains of positive and negative questions are defined as follows:

\begin{itemize}
    \item Positive Questions ($\mathcal{P}$): These are questions for which a positive answer is considered rewarding. For example, questions related to the realism or presence of certain elements in the image can be classified as positive questions.
    
    \item Negative Questions ($\mathcal{N}$): These are questions for which a positive answer incurs a penalty. Questions assessing the presence of unrealistic or undesirable features in the image fall into this category.
\end{itemize}

The assistant provides binary answers ('yes' or 'no') to each question, and the scoring mechanism quantifies the overall quality of the image based on these answers.

%The total score $T_{i}$ for each image is the sum of the individual question scores:

%\begin{equation}
%    T_{i}(I_{j}^{N}) = \sum_{k=1}^{q} S_{i}(I_{j}^{N})
%\end{equation}

\subsection{Open vocabulary object detection}

Based on the provided scores, one or multiple sets of images are subsequently extracted. Out of the remaining image, the following step in the self-rewarding mechanism correspond to the extraction of the best corresponding image through the use of open-vocabulary pre-trained object detection models.

\begin{enumerate}
    \item \textbf{Model Configuration:} The object detection model is configured to recognize objects belonging to the initially specified class. This ensures that the object detection process focuses on features relevant to the class of interest.

    \item \textbf{Iteration through Extracted Images:} For each image in the set of the remaining ones, the object detection model is applied to identify objects within the image. The goal is to locate instances of the specified class within the images.

    \item \textbf{Evaluation of Confidence Values:} The confidence values associated with each detected object are evaluated. The image index (\textit{i}) with the highest confidence value is determined.

    \item \textbf{Extraction of the Optimal Image:} The image for which the detection confidence has been estimated as the highest one is considered as the optimal to the prompt.
\end{enumerate}

The use of open-vocabulary object detection in this context enhances the self-rewarding mechanism by automating the selection of the image that demonstrates the most accurate and confident detection of objects belonging to the specified class.

\subsection{Training Text-to-Image model}\label{sec:fine-tuning-t2i}
%Following the generation of optimal images aligned with specific prompts, the subsequent stage involves training the Text-to-Image (T2I) model on the defined optimal pairs. 
LoRA [\cite{hu2021lora}], short for Low-Rank Adaptation of Large Language Models, presents a pioneering approach devised by Microsoft researchers to efficiently fine-tune expansive language models. LoRA suggests a method wherein the pre-trained language model weights remain fixed, and trainable layers (referred to as rank-decomposition matrices) are introduced within each transformer block. This strategy significantly diminishes the count of trainable parameters and the GPU memory demands during training or inference, as gradients do not need to be computed for the majority of the model weights. By concentrating on the Transformer attention blocks of large-language models, the quality of fine-tuning achieved with LoRA matches that of full model fine-tuning, while being markedly swifter and less resource-intensive. When using LoRA, it is possible to avoid catastrophic forgetting of the text-to-image model abilities while outperforming plain fine-tuning
\cite{kumari2023multi}.
%Textual Inversion [\cite{gal2022image}] is another popular method that attempts to refine a trained Stable Diffusion Model. One of the main reasons for using Textual Inversion is that trained weights are also small and easy to share. However, they only work for a single subject, whereas LoRA can be used for general-purpose fine-tuning, meaning that it can be adapted to new domains or datasets.

%Lora [\cite{hu2021lora}] leverages the low-rank properties of domain adaptation, reducing the number of parameters to train.
LoRA differs from full-parameter fine-tuning in two fundamental parts: it tracks changes to weights instead of updating the weights entirely, and decomposes the weight matrix into two smaller matrices that contain the trainable parameters. Using decomposed matrices reduces the amount of GPU memory usage. By storing the model’s weights (called weights \textit{freezing} [\cite{hu2021lora}]) in memory during the training process, a lot of computation memory is saved. The model's original weight matrix is left unchanged, only new weight matrices are summed to effectively fine-tune the language model.
After training, the LoRA adapter matrices are then loaded to the retrained SD and later used for inference and evaluation. LoRA is accessed within the Hugging Face Parameter Efficient Fine-Tuning (PEFT) library [\cite{peft}].HuggingFace library offers a convenient trainer for supervised finetuning with seamless integration for LoRA training and inference.

%%%%%%%%%%%%%%%%%%%%%%%%%%%%%%%%%%%%%%%%%%
\section{Experiments}\label{sec:Experimentalstudy}

\subsection{Setup configuration}

The implementation has been conducted on an Ubuntu linux 22.04 server with 3 GPUs A6000 each with 48 GB of RAM. To evaluate the proposed self-rewarding T2I approach, the selected models are: (1) the LLM used for the generation of prompts is the AWQ quantized version of Mistral-7B-instruct. (2) the T2I Model selected is Stable Diffusion 2.1. (3) I2T model for captionning is Large language and vision assistant (LlaVa 1.6). (4) the open vocabulary object detection used is YOLO-World. % in Table \ref{tab:model_selection}.

%\begin{table}[ht]
%\centering
%\renewcommand{\arraystretch}{1.0} 
%\caption{\label{tab:model_selection}Summary of models selection}
%\begin{tabular}{lll}
%    \hline
%    Model & Release & Description \\ \hline
%    LLM & Mistral-7B-instruct AWQ & Quantized 7 billion parameters model \\
%    T2I Model & Stable Diffusion 2.1 & Latest version of the generative T2I \\
%    I2T Model & LlaVa & Large language and vision assistant 1.6 \\
%    Object Det. & YOLO-World & Open Vocabulary object detection model\\
%    Text Simil. & Sentence Transformers & Semantic Textual Similarity framework\\
%    \hline
%\end{tabular}
%\end{table}

For prompts generation, the use of a light LLM is sufficient as the goal is to generate a set of prompts that are going to be input to the T2I model. The LLM could possibly be fine-tuned on a set of diffusion models prompts to increase the accuracy of the generated prompts. In the experiment, a set of diffusion prompts templates are given to Mistral 7B-Instruct and asked to generate similar prompts. For the choice of the T2I models, stable diffusion stands as one of the best open sourced image diffusers. The latest version 2.1 allows simple fine-tuning using accelerate library. For the I2T model, the LlaVa assistant is chosen as for it capacity to detail its analysis of an image. Finally, to evaluate the similarity between initially generated prompts and lately obtained image captions.

\subsection{LLM Prompts generation}
\begin{figure}[ht]
	\centering
        \includegraphics[width=\textwidth,scale=1]{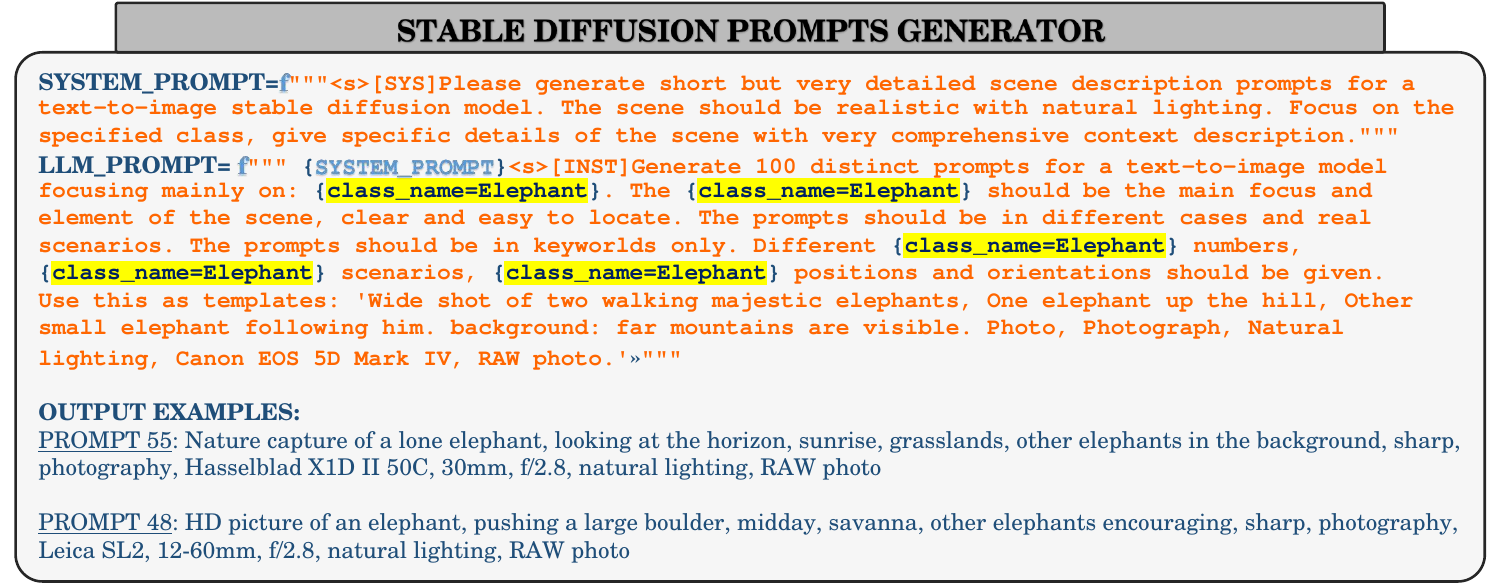}
	\caption{LLM prompting instructions and examples of the generated T2I prompts}
	\label{fig:examples_llm_prompting}
\end{figure}

For this evaluation, the LLM-based generation of diffusion model prompts is conditioned by a set of object classes. Those classes belong to the common objects usually used to train implemented object detection models. We focus on the generation of realistic images of two classes: 'Elephant' and 'Giraffe'. For diffusion model, one way to make the model optimal in following instruction is providing the prompt in the form of multiple keywords describing the desired scenario. This prompt, including SYSTEM PROMPT and an example of the generated T2I prompt is given in Figure \ref{fig:examples_llm_prompting}.

The main followed logic here is to generate photo-realistic image in order to make the next fine-tuned model more efficient in generating photo-realistic pictures of wild animals. For each class, 100 prompts have been sampled with temperature $T$ = 0.7 and probability cutoff for token selection $p$ = 0.95, leading to a total of 200 prompts for both selected classes 'Elephant' and 'Giraffe'.

\subsection{Text-to-Multi Images generation}
This step consists of the generation of $N = 10$ candidate images for each prompt. The images are group in a grid of 2 rows and 5 columns to make the processing less demanding. The selected model is the latest stable diffusion version 2.1 mainly known for its open source use and availability of multiple pre-trained models. Originally, each generated image has a resolution of $512 \times 512$. Therefore, a total of $(10$ images $\times$ $100$ prompts $) = 2000$ images are generated for both classes.% Figure 

\subsection{Self-Rewarding mechanism}
\subsubsection{Self-judging Image-to-Text}
To self-judge the quality of the images, we employed I2T for image captioning and analysis. The selected model for this task is Large Language And Vision Assistant (LlaVa) [\cite{liu2023visual}] which have demonstrated it performance and have been proved to be efficient especially for describing and spotting unusual patterns in the images. To make the vision assistant LlaVa an image content-judging, the initially generated prompts from the LLM are given as an input along with the question about each individual image. 

%\begin{figure}[ht]
%	\centering
%        \includegraphics[width=\textwidth,scale=1]{testing_I2T_rewarding.pdf}
%	\caption{Testing rewarding functions for a specific prompt. Where \textbf{SYS\_PR} is an initial system prompt that takes as input the LLM prompt, and prompts 1 to 4 are the tested Q\&A with LlaVa}
%	\label{fig:testing_I2T_rewarding}
%\end{figure}

The primary goal of this granting strategies is to obtain a final assessment based on the I2T model answering a set of questions about the image content, quality, and visual accuracy to the initially given prompt. Each answer to a positive or negative question prompt will award or withdraw 1 point to the final reward function.
Overall, a total of 10 questions have been asked on each image, this amount has been fixed to have a scoring out of 10.

The scoring strategy consists of asking positive and negative questions and attributing either a positive, null or negative score to the given question. For example, two of the relevant questions are: %(appendix \ref{sec:appendix}):
\begin{itemize}
    \item[(\textbf{Q1})] Does this image look realistic, considering lighting, shadows, and reflections?
    \item[(\textbf{Q2})] Are there any subtle, unexpected patterns or behaviors in the image that might not be immediately noticeable?
\end{itemize}

Where Q1 is a positive question and its positive answer grants a positive score whereas Q2 is rather negative and its positive answer retrieves points from the final score. Within the given example, it seems that the stable diffusion model have successfully been able to generate photo-realistic images with very few outliers where the images seems more to be an artistic art/draw.

\subsubsection{Open Vocabulary object detection filtering}
In some cases, the generated images does not provide a clear view of the aimed class. As a filtering process, the selection of the optimal images out of the selected ones goes through an open vocabulary object detection model. For this process, we have selected the latest YOLO version called YOLO-world a real-time open vocabulary model [\cite{cheng2024yolo}].

This filtering step plays an important role in refining the selection of the best-scoring images, ensuring that only those with a high level of clarity and relevance to the desired class are retained for subsequent stages of evaluation and utilization. By setting a higher threshold, the image could be even more optimized, however, a risk of not being able to detect even clear objects is high. Therefore a balance should be set, in our case 0.6 was a good detection confidence threshold.

\subsection{Fine-tuning stable diffusion}
\begin{figure}[h!]
    \centering
    \includegraphics[width=\textwidth,scale=1]{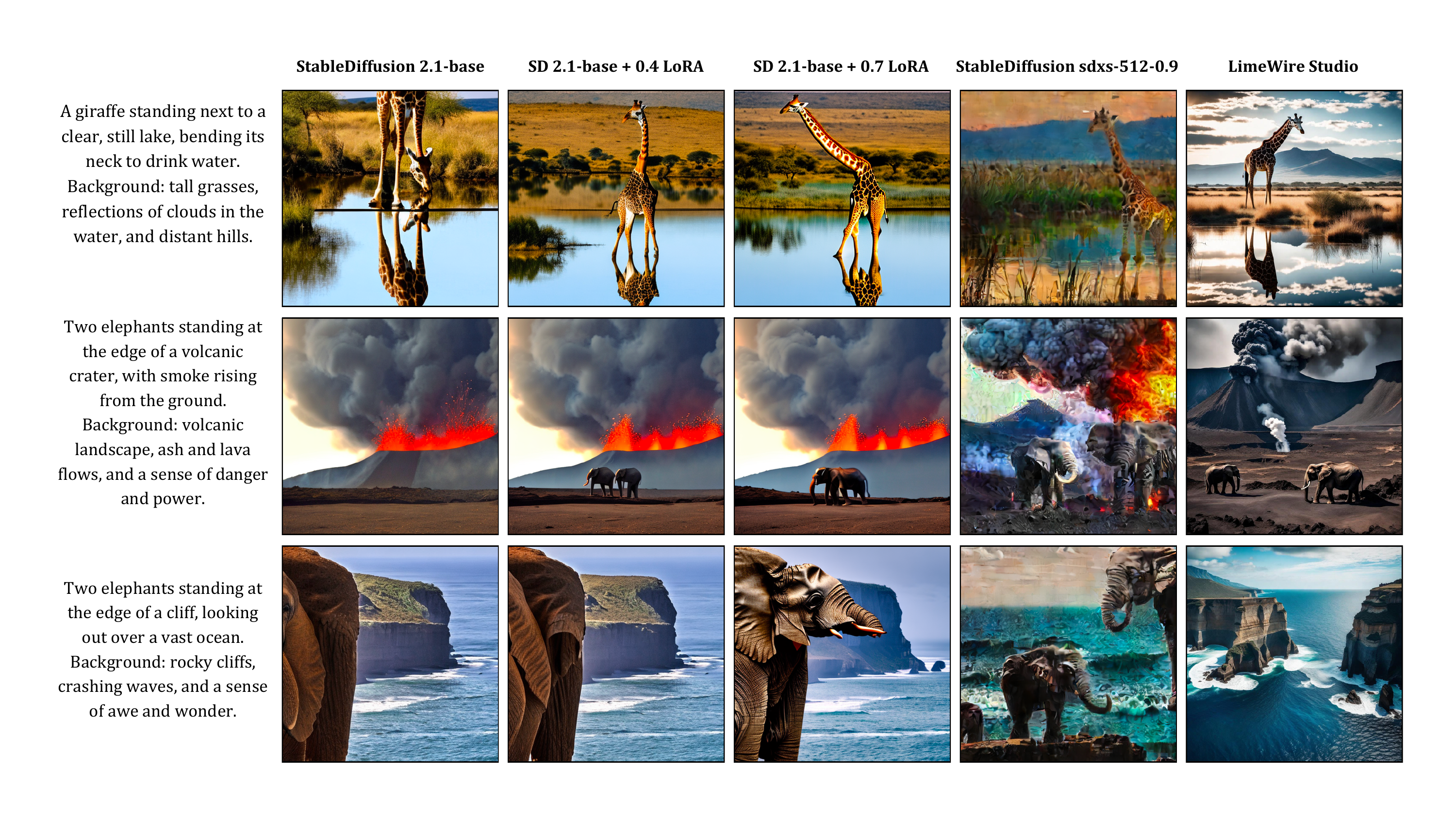}
    \caption{Text-to-image compared results between the base model, our retrained StableDiffusion version at 0.4 and 0.7 LoRA weight intensity, and LimeWire Studio [\cite{Bluewillow}] for comparison. Below the generated images the CLIP score measures the compatibility of image-prompt pairs.}
    \label{fig:fine-tuned_result}
\end{figure}

When fine-tuning stable diffusion, it is necessary to preserve the visual features and general text-to-image capabilities of the base model on any prompt. However, in precise and retrained tasks, the goal of fine-tuning is to increase the precision and perceived accuracy of the model, while preserving the key stable diffusion abilities in text-to-image precision. For these reasons we use LoRA [\cite{hu2021lora}] training as mentioned in \ref{sec:fine-tuning-t2i}.
%Naïvely fine-tuning entire parameters of a pre-trained model with limited target data would be astraightforward approach, but that this is prone to overfitting.
Using the filtered images and prompts pairs, the Stable Diffusion 2.1 [\cite{Rombach_2022_CVPR}] is retrained with LoRA in a traditional supervised fine-tuning fashion. The entire model retraining is done with the Diffusers library of HuggingFace [\cite{von-platen-etal-2022-diffusers}]. The training resolution for the images is set at 512, as this is the resolution of the original StableDiffusion 2.1, and the images are augmented with random horizontal flip. 
The training is completed on 2 NVIDIA A6000 48GB GPUs using mixed precision 16 float format, which allows memory optimization while ensuring no task-specific accuracy is lost compared to full precision training. The model is trained using LoRA for 100 epochs, with a batch size of 18, which is completed in two hours on the GPU hardware. The learning speed is set at a constant rate of 1e-04. 
Due to the nature of LoRA training and the selective dataset we are using, the inference results when applying the LoRA weights may vary. To evaluate the performances and changes, the additional weights are scaled between 0.0 and 1.0, translating to the intensity at which the LoRA weights are loaded. The most interesting changes occur between 0.2 and 0.8 LoRA weight scale, where the weights impact sufficiently the diffusion process without changing and losing the StableDiffusion creation entirely. 
Results of the inference at the LoRA scales 0.2, 0.4 and 0.7 are displayed in Figure \ref{fig:fine-tuned_result}. 
The relationship between prompts and generated images are evaluated using the CLIP score [\cite{radford2021learning}], which provides a qualitative evaluation of the models, necessary for automation. 

\begin{figure}[h!]
    \centering
    \includegraphics[width=\textwidth,scale=1]{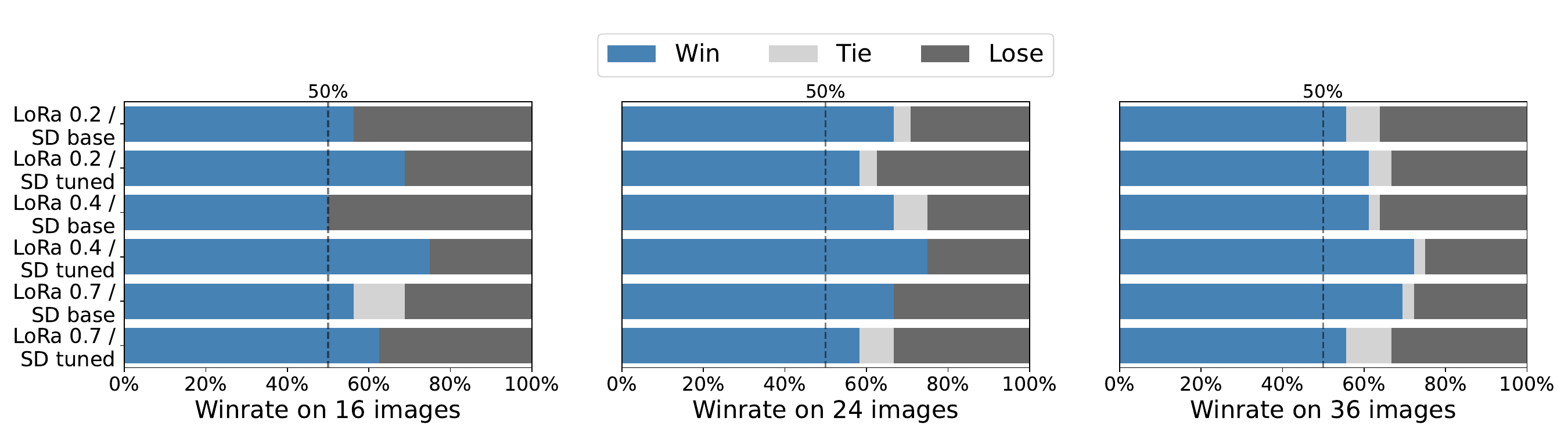}
    \caption{The LoRA trained type models scores are subtracted to the base and fine-tuned models result for each prompt-score pair. A positive CLIP score opposition result is considered a win, and a negative result is a loss. A tie happens when the difference between the scores is less than $\pm$ 0.01. Each of the 50 validation prompts is tested 4(9) times with a different fixed seed.}
    \label{fig:win-rate_result}
\end{figure}

The fine-tuned stable diffusion mode has delivered better CLIP scores also with a more realistic rendering compared to LimeWire. Visually, the fine-tuned model also delivers more accurate and shows an optimized prompt instruction following

\section{Discussion}

\begin{figure}[t]
    \centering
    \includegraphics[width=\textwidth,scale=1]{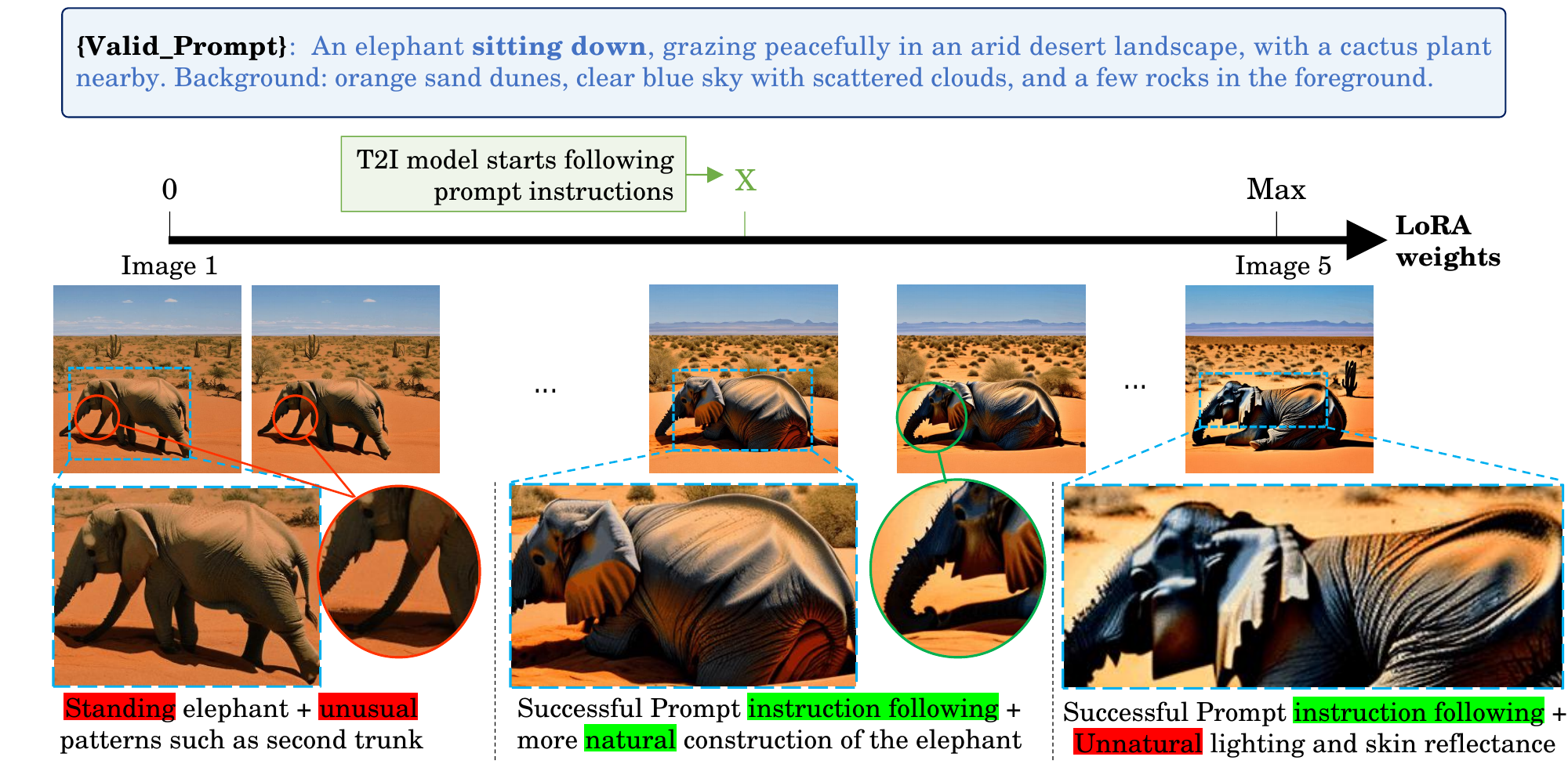}
    \caption{Validation prompt example showcasing instruction following capabilities of the proposed self-rewarding mechanism. Image 1 is the original stable diffusion and after LoRA fine-tuning on self-generated dataset}
    \label{fig:example_Lora_Finetune}
\end{figure}

To evaluate the efficiency of the proposed method, 50 validation prompts are generated using Mistral 7B. The prompts are similar to those generated in the first step of the T2I self-rewarding pipeline. Figure \ref{fig:example_Lora_Finetune} illustrates an example of resulting LoRA-based fine-tuning on the obtained optimal pairs. The first image generated by the base stable diffusion model shows an inaccurate construction as the image does not follow correctly the instructions within the prompt. When fine-tuned, the model adapts better and provides more accurate image. However, training the model with more epochs might lead to overfitting and therefore some unusual patterns and image behaviour as shown in Image 5. Moreover, the base Stable Diffusion model may omit complete precision of the prompt, as seen in Fig. \ref{fig:fine-tuned_result} where the \textit{two elephants} from the very beginning of the prompt are absent on the image. This translates to poorer CLIP scoring as compared to our LoRA-augmented retraining. 

To further evaluate the method's efficiency in enhancing T2I models using validation prompts, the final model was assessed against the baseline SD [\cite{Rombach_2022_CVPR}] and SDXS [\cite{song2024sdxs}] (a fine-tuned version of SD). Each validation prompt was tested using a consistent PyTorch pseudo-random generation seed to ensure reproducibility. Our LoRA weights were adjusted across three different scales (0.2, 0.4, and 0.7) to examine their impacts on the SD checkpoints.
The inference results were qualitatively evaluated using CLIP scoring [\cite{radford2021learning}], similar to the LoRA training phase. The win-rate results for the 50 validation prompts, each tested with four different generation seeds, are illustrated in Fig. \ref{fig:win-rate_result}. In nearly all comparisons (with one exception pending results), our method demonstrated a higher win percentage against both the base and the SDXS models. Despite being close to 50\%, the model frequently performed at least as well (wins + ties) as the baseline and fine-tuned SD, achieving an overall win-rate percentage of 70\%.

\textbf{Limitation}

While the proposed class-conditional self-rewarding mechanism highly performs compared to other generally trained ones, without the need of human feedback, it still has few limitations. The class-conditional term means that those performances are limited by the specificity of the object images to be generated. This constraint inherently restricts the versatility and generalizability of the model, as it relies heavily on the adequacy of the conditioned classes to cover the breadth of potential image-text relationships.
One solution to this limitation would be continuously performing the CCSR loop while not limiting to few classes, which will lead to an enhanced overall general model. Hence, while the proposed CCSR framework represents a promising advancement in the realm of text-to-image modeling, it is crucial to acknowledge and address these limitations to ensure its robustness and applicability across various domains and contexts.

\section{Conclusion}
CCSR is a self-rewarding mechanism that's proposed to self-improve text-to-image models through the use of image captioning methods. This mechanism in inspired from language models and can replace the traditional fine-tuning methods that are mainly based on reinforcement learning conducted on human preference data. The proposed model consists of 6 steps: (1) LLM prompts generation. (2) Text-to-Image generation. (3) Self-judging using Image-to-Text. (4) Image filtering. (5) Optimal pairs extraction. And (6) Fine-tuning. This framework allows a complete automation and continuous improvement of Text-to-Image models without the need of human intervention. To assess the effectiveness of the proposed mechanism, an experimental test has been conducted through the fine-tuning of stable diffusion on images generated based on Mistral-7B delivered prompts. The overall result in comparison to other models have shown a better visual concordance with the text while also obtaining an improved visual content along with high scoring compared to the pre-trained stable diffusion and another fine-tuned model namely SDXS.

\section{Conclusion}
CCSR is a self-rewarding mechanism that's proposed to self-improve text-to-image models through the use of image captioning methods. This mechanism in inspired from language models and can replace the traditional fine-tuning methods that are mainly based on reinforcement learning conducted on human preference data. The proposed model consists of 6 steps: (1) LLM prompts generation. (2) Text-to-Image generation. (3) Self-judging using Image-to-Text. (4) Image filtering. (5) Optimal pairs extraction. And (6) Fine-tuning. This framework allows a complete automation and continuous improvement of Text-to-Image models without the need of human intervention. To assess the effectiveness of the proposed mechanism, an experimental test has been conducted through the fine-tuning of stable diffusion on images generated based on Mistral-7B delivered prompts. The overall result in comparison to other models have shown a better visual concordance with the text while also obtaining an improved visual content along with high scoring compared to the pre-trained stable diffusion and another fine-tuned model namely SDXS.

%\section*{Acknowledgments}
%This was was supported in part by......

%Bibliography
\bibliographystyle{elsarticle-harv}  
\bibliography{references}  

%%%%%%%%%%%%%%%%%%%%%%%%%
%%%%%%%  APPENDIX %%%%%%%
%%%%%%%%%%%%%%%%%%%%%%%%%
\newpage
\appendix
\onecolumn
\section*{APPENDIX}\label{sec:appendix}

\begin{figure}[htb]
	\centering
        \includegraphics[width=\textwidth,scale=1]{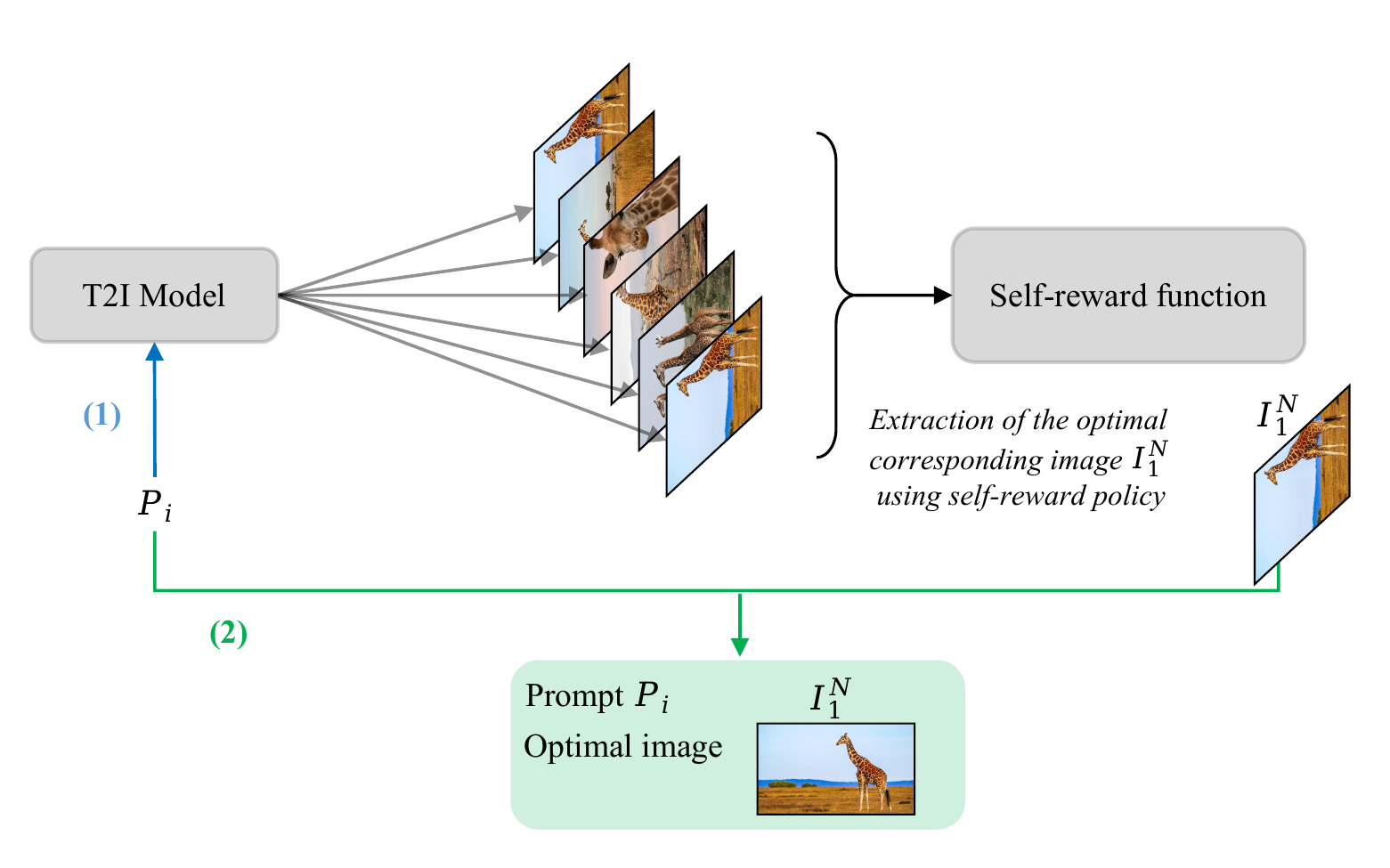}
	\caption{visualization of the self-reward policy function from multi-images generation. Where (1) is the first path for image generation and (2) is the second path of optimal pairs extraction}
	\label{fig:self_reward_from_multi-images}
\end{figure}

For a given prompt $P_i$, multiple images are generated without seeding the models in other to have room for innovation. This first step is represented by the path (1) in the Figure \ref{fig:self_reward_from_multi-images}. Next, the Self-reward function (path (2) in Figure \ref{fig:self_reward_from_multi-images}) has as task to check the generated images and output the best scoring one. The extracted images $I$ along yith the initial prompt is the considered as the optimal pair.

\begin{figure}[htb]
	\centering
        \includegraphics[width=\textwidth,scale=1]{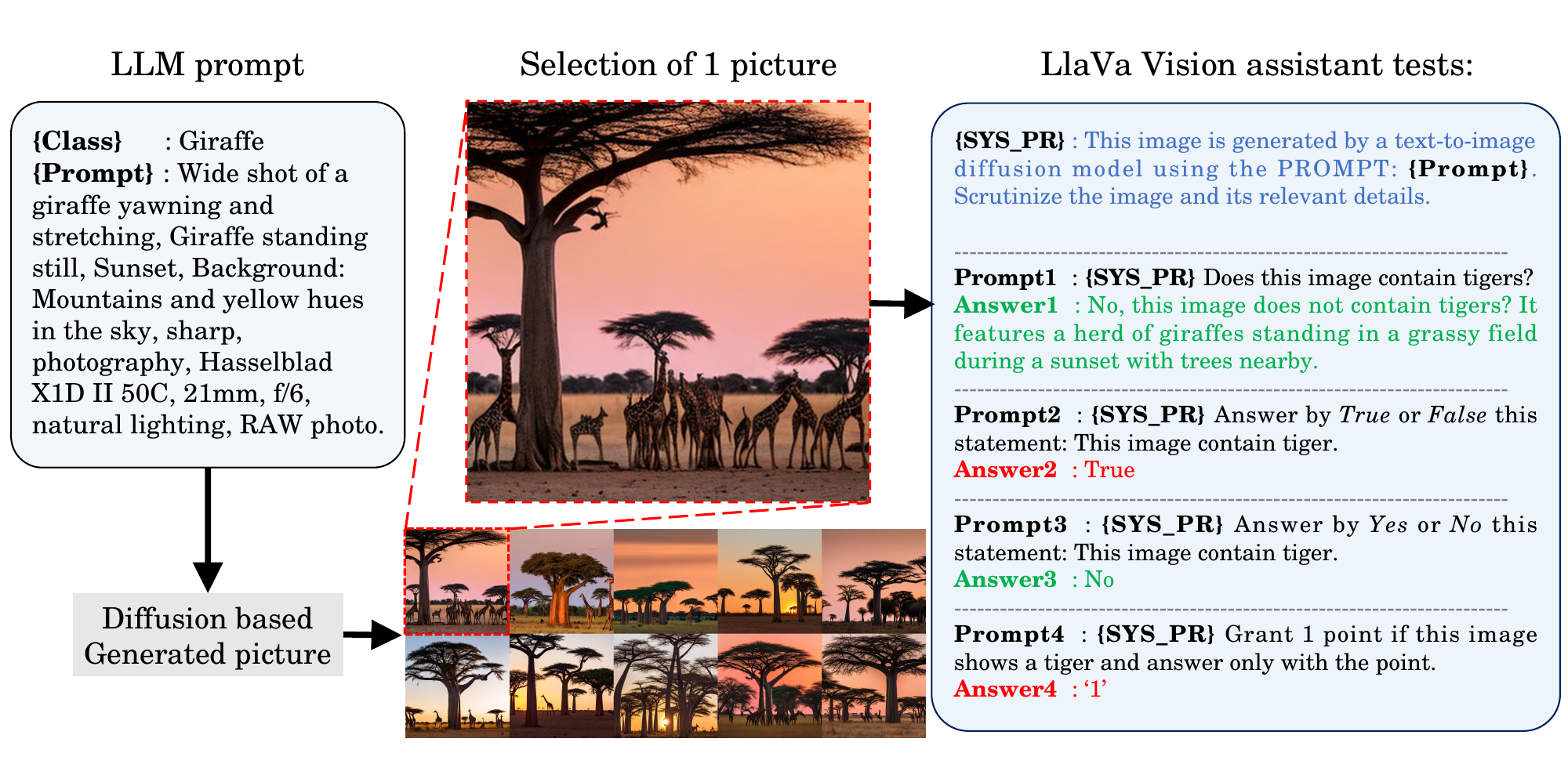}
	\caption{Testing rewarding functions for a specific prompt. Where \textbf{SYS\_PR} is an initial system prompt that takes as input the LLM prompt, and prompts 1 to 4 are the tested Q\&A with LlaVa}
	\label{fig:testing_I2T_rewarding}
\end{figure}

To evaluate various granting strategies within the LlaVa model, we experimented with different prompts aimed at determining the presence and absence of an object in a given image. Notably, in the Figure \ref{fig:testing_I2T_rewarding}, both prompts 1 and 3 yielded valid responses, where the prompt 1 accurately detailed the description of image content. However, prompts 2 and 4 did not produce reliable results, showcasing a disparity in the comprehensibility of the prompting strategies. These findings emphasize the importance of crafting prompts that align well with the model's capabilities for effective image analysis and content recognition.

While applying the rewarding strategy using the prompt3 in the previous example, the best scoring images are 5, 7, 8 and 10. Those images are the best at following the user prompt instruction (Table \ref{tab:image_scores}).

\begin{table}[H]
\renewcommand{\arraystretch}{1.3} 
\centering
\caption{\label{tab:image_scores}Self-judging scoring strategy example of the 10 generated images for a single prompt while answering 10 distinct questions about the image quality and content relevance to the given initial prompt}
\begin{adjustbox}{width=370 pt}
\begin{tabular}{l*{10}{c}c}
    \toprule
     & Q1 & Q2 & Q3 & Q4 & Q5 & Q6 & Q7 & Q8 & Q9 & Q10 & Total Score \\
    \midrule
    \rowcolor{gray!10}Image 1 & 1 & -1 & 1 & 1 & 1 & 0 & 1 & 1 & 1 & 1 & 7 \\
    \rowcolor{gray!10}Image 2 & 1 & -1 & 1 & 1 & 1 & 0 & 1 & 1 & 1 & 1 & 7 \\
    \rowcolor{gray!10}Image 3 & 1 & -1 & 1 & 1 & 1 & 0 & 1 & 1 & 1 & 1 & 7 \\
    \rowcolor{gray!10}Image 4 & 1 & 0 & 0 & 0 & 0 & 0 & 1 & 1 & 1 & 1 & 5 \\
    \rowcolor{green!20}\textbf{Image 5} & 1 & 0 & 1 & 1 & 1 & 0 & 1 & 1 & 1 & 1 & \textbf{8} \\
    \rowcolor{gray!10}Image 6 & 1 & -1 & 1 & 1 & 1 & 0 & 1 & 1 & 1 & 1 & 7 \\
    \rowcolor{green!20}\textbf{Image 7} & 1 & 0 & 1 & 1 & 1 & 0 & 1 & 1 & 1 & 1 & \textbf{8} \\
    \rowcolor{green!20}\textbf{Image 8} & 1 & 0 & 1 & 1 & 1 & 0 & 1 & 1 & 1 & 1 & \textbf{8} \\
    \rowcolor{gray!10}Image 9 & 1 & -1 & 1 & 1 & 1 & 0 & 1 & 1 & 1 & 1 & 7 \\
    \rowcolor{green!20}\textbf{Image 10} & 1 & 0 & 1 & 1 & 1 & 0 & 1 & 1 & 1 & 1 & \textbf{8} \\
    \bottomrule
\end{tabular}
\end{adjustbox}
\end{table}
where the list of asked questions are:
\begin{itemize}
    \item[\textbf{(Q1)}] Does this image look realistic, considering lighting, shadows, and reflections?
    \item[\textbf{(Q2)}] Are there any subtle, unexpected patterns or behaviors in the image that might not be immediately noticeable?
    \item[\textbf{(Q3)}] can you clearly see \{$class\_name$\} in the image?
    \item[\textbf{(Q4)}] (if the answer of the previous question is 'No' answer 'Nan' to this question) Considering specific details like \{$class\_name$\} posture, head and body shape, number of legs and form and the surroundings view, does the image look normal?
    \item[\textbf{(Q5)}] (if response to 3. is 'No' answer 'Nan' to this question) If \{$class\_name$\} is present, does it exhibit realistic and natural behavior?
    \item[\textbf{(Q6)}] (if response to 3. is 'No' answer 'Nan' to this question) Are there any other objects or elements in the image that might be mistakenly identified as \{$class\_name$\}?
    \item[\textbf{(Q7)}] (if response to 3. is 'No' answer 'Nan' to this question) Does the representation of \{$class\_name$\} in the image maintain anatomical accuracy (e.g., correct number of legs, tail, head)?
    \item[\textbf{(Q8)}] Do the colors in the image accurately match the description in the PROMPT \{$prompt3$\}, including subtle variations and shades?
    \item[\textbf{(Q9)}] Can you identify any deviations or abnormalities in the image that might be less obvious but still significant?
    \item[\textbf{(Q10)}] Given PROMPT: \{$prompt3$\}. Does the image respect fully the prompt description?
\end{itemize}
    
\begin{figure}[H]
\centering
        \includegraphics[width=\textwidth,scale=1]{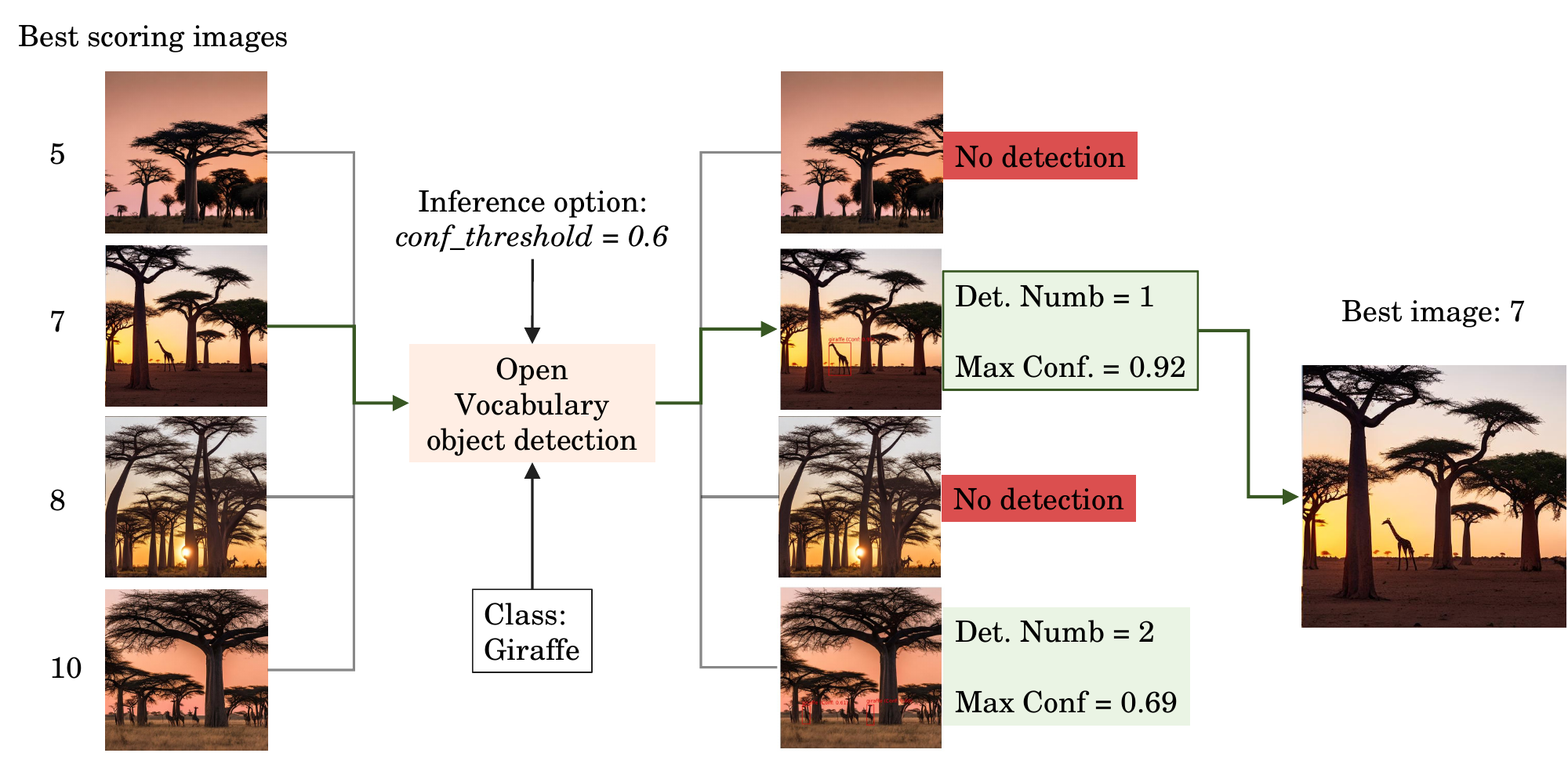}
	\caption{Open vocabulary object detection example: The model is fed with the best scoring images, the name of the class is given as the aimed for detection object, and finally the minimum detection confidence considered is 0.6}
	\label{fig:example_open_vocab_det}
\end{figure}

To filter out of the best scoring image one that's mostly containing the object, we propose the use of open vocabulary object detection algorithm. The objective is to ensure that the object is visible and well recognized by pre-trained detection models. The selection process is summarized in Figure \ref{fig:example_open_vocab_det}.

\begin{figure}[htb]
\centering
        \includegraphics[width=\textwidth,scale=1]{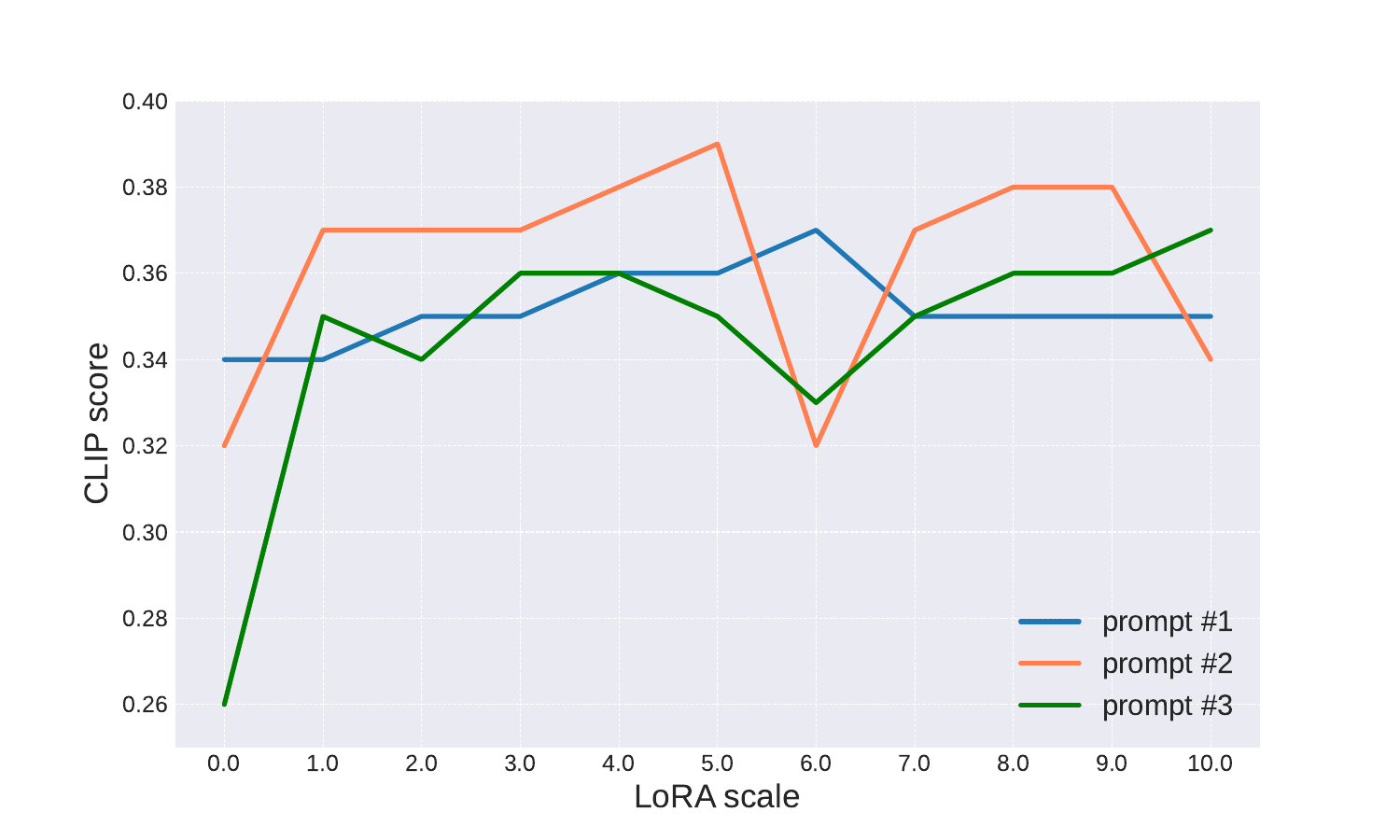}
	\caption{CLIP score curve when applying self-rewarded fine-tuned stable diffusion LORA weights between 0 and 1 range.}
	\label{fig:lora_clip}
\end{figure}

\begin{figure}[htb]
\centering
        \includegraphics[width=\textwidth,scale=1]{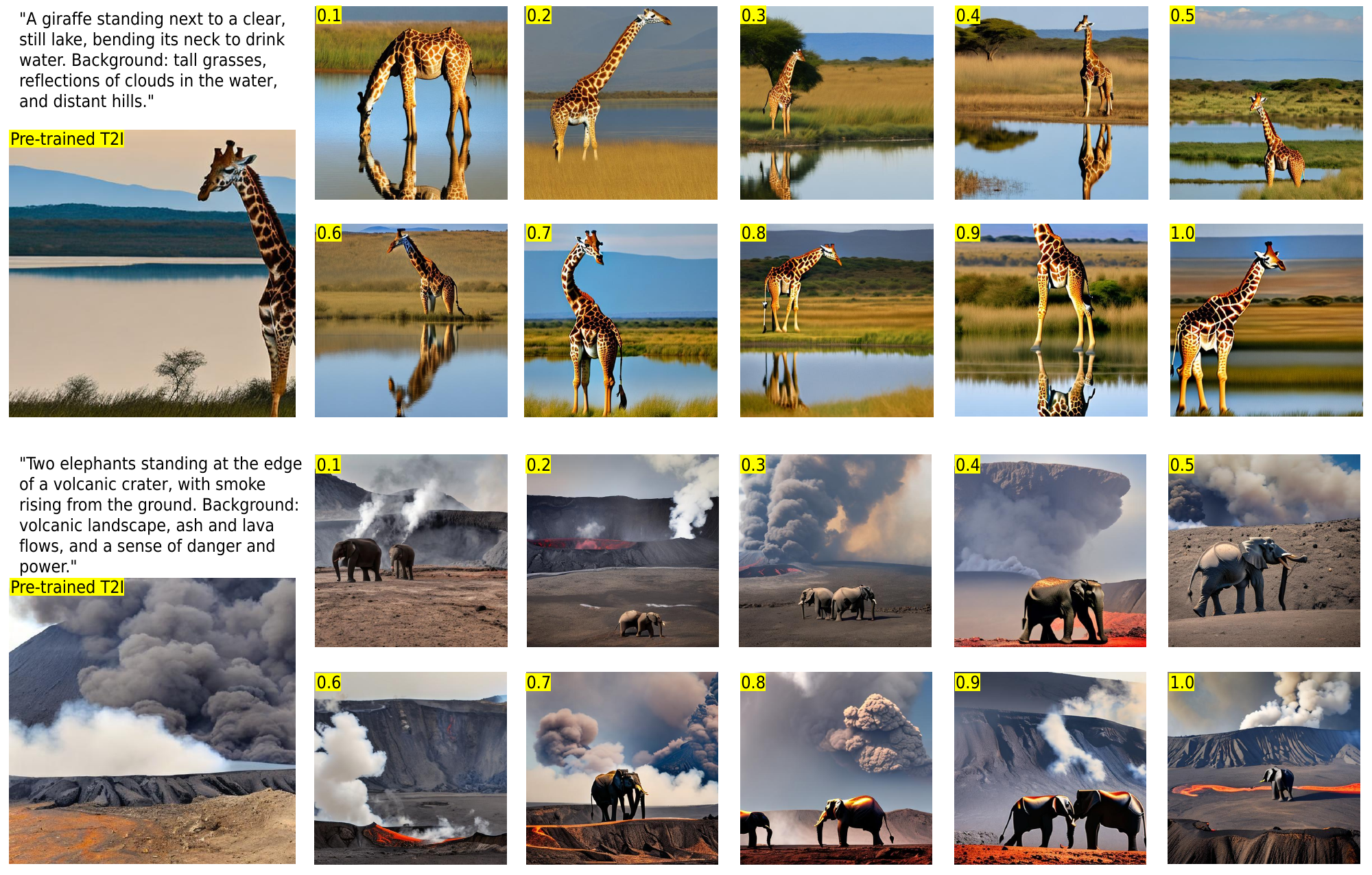}
	\caption{Visualization of image generation variation with LORA weights fine-tuning.}
	\label{fig:lora0to1}
\end{figure}

Once the optimal pairs are extracted, the Stable diffusion has been fine-tuned based on the dataset. To assess the quality of the obtained images compared to pre-trained stable diffusion, a comparative study based on CLIP score has been conducted. Figure \ref{fig:lora_clip} shows the evolution curves of 3 validation prompts from pre-trained (0 LORA scale) to 1 (maximum LORA scales) where we can distinguish an improvement of the scoring when applying the proposed self-rewarding mechanism. Resulting images for 2 example prompts are visualized in Figure \ref{fig:lora0to1} where we can observe an improve instruction following.

%%%%%%%%%%%%%%%%%%%%%%%%%%%%%%%%%

\end{document}